# Toward a Cross-Lingual Romanization Ecosystem for Sinitic Languages: A Paired Mandarin–Cantonese Case Study

*Zijie Zhang*[1], *Tan Lee*[1*], *Yong Cao*[1], *Benyou Wang*[1*]

[1]The Chinese University of Hong Kong, Shenzhen

zijiezhang@link.cuhk.edu.cn, tanlee@cuhk.edu.cn, wangbenyou@cuhk.edu.cn

## Abstract

This paper proposes the Sinitic Romanization Ecosystem, a cross-lingual Sinitic romanization design framework with supporting digital infrastructure and a community-driven open-source workflow. The design framework addresses the lack of systematic cross-lingual romanization alignment among Sinitic languages through four design principles: phonetic correspondence for representing similar sounds with similar romanized symbols, historical-phonological correspondence for aligning cognate romanization strings, one-phoneme-one-symbol, and basic Latin-letter use, with a balancing consideration recognizing trade-offs among these principles. For the main paired case study, we develop CantRomZJ1 and MandRomZJ1, Cantonese and Mandarin romanization schemes following the design framework, respectively. We also develop schemes for several other Sinitic languages, including Meixian Hakka, Shanghai Wu, and Nanjing Jianghuai Mandarin, following the same design framework. To bring the romanization schemes into practical use, we develop open-source infrastructure for structured romanization storage, conversion, parsing, dictionary construction, and input-method generation. Finally, we evaluate the design framework through speech-to-romanization experiments based on Meta's Massively Multilingual Speech (MMS) fine-tuning. Compared with the Pinyin+Jyutping baseline, our MandRomZJ1+CantRomZJ1 condition reduces Cantonese WER and CER by 7.80% and 10.61%, respectively. These results suggest that cross-lingual romanization alignment can improve transfer in low-resource Sinitic speech technology.

## 1. Introduction

Romanization broadly refers to the representation of a language's written form in the Latin script. In this paper, it refers specifically to Latin-script representations of pronunciations in Sinitic languages. Such representation is particularly important because Chinese characters, as a primarily logographic script, do not directly indicate pronunciations across different Sinitic languages [1]. Modern Standard Chinese (hereafter Mandarin unless otherwise specified) and non-Mandarin Sinitic languages, traditionally referred to as Chinese dialects, all belong to the Sinitic language family. They are closely related and share many cognates. Most non-Mandarin Sinitic languages do not have a standardized writing system. Nevertheless, they have readings for a large number of Chinese characters, which function as written representations of cognates across the language family [1]. For example, Cantonese (the Yue Chinese language spoken in urban Guangzhou, Hong Kong, and Macau; Yue Chinese is a branch of Sinitic languages mainly distributed in Guangdong, Hong Kong, Macau, and Guangxi; Cantonese is a representative Yue Chinese language) has readings for more than 20,000 Chinese characters [1], [2]. Some of these Chinese characters, however, do not correspond to commonly used morphemes in spoken Cantonese. Therefore, romanization systems are needed to represent the pronunciations of Chinese characters in different Sinitic languages more directly.

Communities, scholars, and institutions have developed romanization schemes for various Sinitic languages and Sinitic branches or groups, traditionally referred to as dialect groups (hereafter language groups) [3]–[8]. These schemes are carefully optimized for the target languages or language groups. Within a group, shared conventions or alignment among related languages are maintained. This suggests that alignment is an important practical concern in Sinitic language romanization. However, such alignment is usually limited to a single language group and has not been systematically extended across different language groups. As a result, similar sounds may be represented by distinct romanized symbols, and cognates may lose spelling similarity across language groups. From the perspective of human learning, such divergence may increase the learning burden for users [9]–[11]. From the perspective of language technology, reduced romanization string similarity may impact cross-lingual transfer in low-resource settings [12], [13].

Beyond romanization schemes, digital infrastructure is crucial for bringing romanization from theoretical design into practical use. Existing schemes have been digitized and developed as Chinese-character input methods, online pronunciation lookup, and other user-facing tools [2], [3], [6], [14], [15]. Jyutping is a well-established Cantonese romanization scheme developed by the Linguistic Society of Hong Kong (LSHK) [7]. It is an exemplary case of romanization digitization with high-quality data and software support for grapheme-to-phoneme (G2P), syllable parsing, and NLP processing [2], [16]–[18]. These practices show that data interoperability is crucial. Files storing romanization schemes structurally can support syllable parsing, teaching, and pronunciation-dictionary construction. Here, a pronunciation dictionary refers to a structured mapping between Chinese characters and their romanized pronunciations. Such pronunciation dictionary files can support lookup, input methods, grapheme-to-phoneme (G2P), and phoneme-to-grapheme (P2G). However, existing schemes and pronunciation dictionaries are stored and displayed in heterogeneous formats, making it difficult to develop reusable software for cross-lingual data analysis [2]–[8], [14], [18].

To address these issues, this paper proposes the Sinitic Romanization Ecosystem, which consists of a romanization design framework, the digitization infrastructure, and an open-source resource-sharing workflow. The design framework stipulates the principles and a consideration in designing romanization schemes for Sinitic languages. Among the principles, there are two correspondence principles: phonetic correspondence and historical-phonological correspondence. Phonetic correspondence aligns similar sounds across languages. Historical-phonological correspondence preserves spelling similarity among cognates. There are also two practical principles: one-phoneme-one-symbol and basic Latin-letter use [19]. The design framework also includes a balancing consideration, recognizing that these principles may involve trade-offs in actual romanization design. We present CantRomZJ1 and MandRomZJ1, Cantonese and Mandarin romanization schemes respectively, as the main paired case study of the design framework. We also apply the design framework to develop romanization schemes for a few other Sinitic languages, including Meixian Hakka, Shanghai Wu, and Nanjing Jianghuai Mandarin.

With the infrastructure, the romanization schemes are brought into practical use as reusable resources. PhonEngine comprises PhonSymbol, providing a unified romanization-scheme entry interface and unified structured storage, and PhonConvert, providing efficient large-scale pronunciation-dictionary construction. For Cantonese, we provide CantRomZJ1 data and tools linked to existing Jyutping resources, which are also distributed via PyPI as PySinRom (package name: pysinrom) [16]–[18]. There is also a tool for converting pronunciation dictionaries into Rime input method engine-compatible YAML files for input-method deployment [15], [20]. All these infrastructure resources are publicly available as open-source resources in our GitHub repository: https://github.com/Zijie-ZHANG-Ling/Sinitic-Romanization-Ecosystem. The repository supports open-source development by allowing users to share, fork, extend, and contribute PhonEngine-format romanization data for more languages. This implementation establishes the Sinitic Romanization Ecosystem as a community-maintained resource base that can grow beyond the languages presented in this paper.

A preliminary ASR-based validation is also carried out by fine-tuning Meta's Massively Multilingual Speech (MMS) model [21]. The results show that, compared with using Jyutping and Pinyin as romanization targets for Cantonese and Mandarin respectively [7], using CantRomZJ1 and MandRomZJ1 improves low-resource Cantonese speech-to-romanization (S2R) recognition when jointly trained with higher-resource Mandarin data.

## 2. Romanization Design Framework

The core idea of the romanization design framework is cross-lingual romanization alignment for both human learning and computational transfer. For human users, alignment allows learners to reuse familiar script–sound knowledge and perceived cognate similarity when learning a new romanization, consistent with findings from psychology studies [9]–[11]. For computational models, alignment can help in two related ways. First, it improves explicit string similarity for lookup, dictionary matching, transliteration, and P2G. For example, a CantRomZJ1 form ⟨jen2seng1⟩ for 人生 is closer to Pinyin/MandRomZJ1 ⟨ren2sheng1⟩ than Jyutping ⟨jan6sang1⟩ is, which may help infer the corresponding Chinese characters when a pronunciation is absent from a given dictionary [12]. Second, it makes romanization data distributions and patterns more similar across related languages. This supports parameter sharing and transfer learning by allowing shared models to transfer acoustic-to-symbol or text representations more easily from a higher-resource system to a lower-resource target system [13]. This design framework is particularly relevant to Sinitic languages because they combine strong relatedness with highly uneven resource distribution: systems such as Mandarin or, regionally, Cantonese are much more resource-rich than many related languages. Therefore, in this design framework, a target language romanization may be designed with reference to a higher-resource system, such as Mandarin in the Chinese context, or to another locally dominant or infrastructurally mature language, such as Cantonese in Lingnan, when appropriate. What matters is not mechanical imitation of a single reference system, but systematic correspondence between the target and reference systems.

The International Phonetic Alphabet (IPA) is a standard phonetic transcription system and is valuable for linguistic precision [22]. However, the design framework targets romanization because it is more compatible with users' established practices such as the input methods and lookup tools. Romanization can also reduce label sparsity and improve parameter sharing in low-resource modeling. For example, representing a sparsely attested [ɬ] in a low-resource language with a romanized ⟨l⟩ can allow models to exploit similarity with /l/ in related languages.

We define two correspondence principles: phonetic correspondence and historical-phonological correspondence, and two practical principles: one-phoneme-one-symbol and basic Latin-letter use. The two practical principles are motivated by the Jyutping scheme design [19].

### 2.1. Phonetic Correspondence

**Phonetic correspondence** means that phonemes with similar or identical phonetic realization across languages should be represented by similar or identical romanized symbols. For example, when Pinyin/MandRomZJ1 is used as the reference system, an unaspirated alveolar affricate /ts/ is preferably represented by ⟨z⟩, and an /ə/-like vowel is preferably represented by ⟨e⟩. Such choices allow human users and computational models to transfer the existing knowledge of high-resource Pinyin/MandRomZJ1. For language learners, this is consistent with Odlin's account that shared script–sound knowledge between an existing familiar system and a target one can facilitate the reuse of prior knowledge and accelerate target-system learning [9].

### 2.2. Historical-Phonological Correspondence

**Historical-phonological correspondence** means that cognates should preserve spelling similarity in their romanized forms across languages. Such similarity would make historical relatedness among Sinitic languages more perceivable to users and improve cross-lingual romanization alignment for computational use. This principle is consistent with Kellerman's account of perceived similarity: when learners perceive forms as related, they are more likely to transfer prior knowledge from an existing familiar system to a target one [10], [11].

### 2.3. Practical Principles

Following the Jyutping scheme design, we adopt the one-phoneme-one-symbol principle, which requires each phoneme to be represented consistently and independently of its surrounding spelling context. The basic Latin-letter use principle follows Jyutping in avoiding diacritics and non-ASCII characters, since basic Latin letters and digits are more desirable for input methods, data storage, software processing, and cross-platform implementation [19].

### 2.4. Tone Representation: Dual Track Design

For initials and finals (e.g., ⟨zh⟩ as initial and ⟨ang⟩ as final in Pinyin syllable ⟨zhang⟩), romanization can usually represent segmental phonemes with Latin letters or letter combinations. Tone, however, is a suprasegmental feature and is difficult to represent directly with basic Latin letters [23]. It therefore requires an explicit notational method. In our design framework, tone representation follows the same two correspondence principles: tone value serves phonetic correspondence and tone category serves historical-phonological correspondence.

Tone value refers to the actual pitch realization of a tone. Tone category refers to its historical source, especially the traditional four categories of the Middle Chinese Qièyùn system: level (píng 平), rising (shǎng 上), departing (qù 去), and entering (rù 入) [1]. Since tone values change rapidly over the historical development of Sinitic languages, the relationship between tone values and tone categories is often not transparent [24]. We therefore adopt a dual-track tone design. For tone value, we use Chao's five-level tone system, one of the most widely used systems for representing relative pitch in Sinitic languages [25], [26]. For tone category, we propose a Hierarchical Tone-Category Notation (HTCN), which encodes historical tone origin through progressively finer levels of classification. The first digit, 1, 2, 3, or 4, represents the level, rising, departing, or entering; the second symbol, A or B, marks a yīn/yáng register split in 1, 2, 3, or 4; and the third symbol, a or b, marks an upper/lower subdivision in entering tones of the Yue Chinese (including Cantonese) when needed [27]. For example, yīn level is represented as 1A, yáng level as 1B, rising as 2, upper yīn entering as 4Aa, and lower yīn entering as 4Ab.

The HTCN makes historical origin clear while allowing language-specific adaptation. For example, rising-tone reflexes across Sinitic languages share the first digit 2, whether they remain unsplit as 2 or split differently into 2A or 2B.

### 2.5. Balancing Between Principles

Given the geographic spread and historical depth of Sinitic languages [1], phonetic correspondence and historical-phonological correspondence may involve trade-offs. Phonemes with similar phonetic realizations across languages may come from different historical origins, while cognates may develop very different pronunciations. Even within a single language, historically related categories may split into sharply different modern forms due to conditioned sound change, language contact, or other factors [1]. Therefore, romanization design requires a balancing consideration to decide when to prioritize which principle. Although such trade-offs may be partly evaluated through measures such as cross-lingual edit distance, learning outcomes, or NLP and speech performance, it is hard to use a single metric to fully determine a universally optimal design. The schemes proposed in this paper therefore represent one possible balance within the design framework, not the only possible solution.

## 3. Paired Case Study: MandRomZJ1 and CantRomZJ1

We introduce CantRomZJ1 and MandRomZJ1 as the main paired case study of Cantonese and Mandarin within the proposed design framework. Because the two languages are closely related but unevenly resourced in language technology, with Mandarin as the higher-resourced one, we design their romanization schemes jointly to improve cross-lingual alignment. Mandarin romanization serves as the primary external reference for Cantonese because Mandarin is one of the most widely learned, resource-rich, and technologically mature Sinitic languages. Conversely, given the influence and technological maturity of Cantonese among non-Mandarin Sinitic languages [27], CantRomZJ1 is also intended to serve as a reference point for neighboring Sinitic languages.

Pinyin is the standard romanization of Mandarin and is widely accepted, which suggests that it should be given strong consideration as a reference system. However, its design is not fully consistent with our principles of one-phoneme-one-symbol and basic Latin-letter use. MandRomZJ1 is therefore developed on the basis of Pinyin, preserving most of its style and conventions while modifying only a limited number of representations to make them consistent with the principles. MandRomZJ1 is not intended to replace Pinyin in general use. It is currently used only in language technology applications and is intended merely as an alternative Mandarin romanization system for interested users.

CantRomZJ1 is developed through modifications to Jyutping [7]. CantRomZJ1 and MandRomZJ1 preserve the phonological systems underlying Jyutping and Pinyin, respectively. We first discuss the modifications from Pinyin to MandRomZJ1, followed by the modifications from Jyutping finals to CantRomZJ1 finals and the dual-track tone representations. CantRomZJ1 retains the same initial romanizations as Jyutping, the rationale for which is presented in the final subsection.

### 3.1. MandRomZJ1 Modifications

**Pinyin ⟨ui⟩/⟨un⟩/⟨iu⟩ → MandRomZJ1 ⟨uei⟩/⟨uen⟩/⟨iou⟩**: This modification is motivated by the one-phoneme-one-symbol principle. The finals ⟨uei⟩, ⟨uen⟩, and ⟨iou⟩ in the official Pinyin scheme are abbreviated as ⟨ui⟩, ⟨un⟩, and ⟨iu⟩ respectively in the real spelling of non-zero initial syllables. Because these finals consist of the medials ⟨u⟩ or ⟨i⟩ followed by ⟨ei⟩, ⟨en⟩, or ⟨ou⟩, MandRomZJ1 retains the unabridged underlying forms in all syllables, changing from the Pinyin abbreviated forms.

**Pinyin ⟨ong⟩/⟨ao⟩/⟨iao⟩ → MandRomZJ1 ⟨ung⟩/⟨au⟩/⟨iau⟩**: This modification is also motivated by the one-phoneme-one-symbol principle. Pinyin ⟨ong⟩ is changed to ⟨ung⟩, and Pinyin ⟨ao⟩ and ⟨iao⟩ are changed to ⟨au⟩ and ⟨iau⟩ respectively. The ⟨o⟩ in these finals actually represents the underlying /u/ phoneme but is written as ⟨o⟩ in Pinyin partly to avoid visual confusion in handwritten forms [28]. This modification reflects the underlying /u/ phoneme more transparently and better follows the one-phoneme-one-symbol principle.

**Pinyin ⟨ü⟩ → MandRomZJ1 ⟨yu⟩**: Pinyin represents /y/ with context-dependent forms: ⟨ü⟩, ⟨u⟩ after ⟨j, q, x⟩, and ⟨yu⟩ in zero-initial syllables [28]. MandRomZJ1 uniformly represents /y/ as ⟨yu⟩ in all contexts and does not add an additional ⟨y⟩ before ⟨yu⟩ in zero-initial syllables. Thus, syllables like /y/ and /yn/ are represented as ⟨yu⟩ and ⟨yun⟩, respectively. This representation follows the one-phoneme-one-symbol and basic Latin-letter use principles. It is also supported by existing conventions: familiar zero-initial form ⟨yu⟩ representing /y/ in Pinyin, ⟨lyu⟩ and ⟨nyu⟩ as transcriptions of Pinyin ⟨lü⟩ and ⟨nü⟩ in Chinese exit-entry documents [29], and ⟨yu⟩ for /y/ in Jyutping and CantRomZJ1.

**Pinyin ⟨xiong⟩/⟨jiong⟩/⟨qiong⟩ → MandRomZJ1 ⟨xyung⟩/⟨jyung⟩/⟨qyung⟩**: Phonologically, the final in ⟨xiong⟩/⟨jiong⟩/⟨qiong⟩ is /yŋ/, but it is represented as ⟨iong⟩ in Pinyin partly to avoid visual confusion in handwritten forms [28]. Following the one-phoneme-one-symbol principle and the ⟨yu⟩ representation discussed in the last paragraph, these three syllables are represented as ⟨xyung⟩, ⟨jyung⟩, and ⟨qyung⟩ respectively.

Apart from these modifications, MandRomZJ1 follows

Pinyin. For other controversial issues of Mandarin phonology, such as whether [ʅ], [ʮ], and [i] should be analyzed as one phoneme [28], we keep the same phonology underlying Pinyin. Following the balancing consideration, these MandRomZJ1 design choices are not the only possible solutions, and representing these phones with different symbols may also be reasonable for certain goals.

### 3.2. CantRomZJ1 Modifications in Finals

CantRomZJ1 modifies the representation of several nucleus phonemes in Jyutping. Specifically, the Jyutping ⟨aa⟩, which represents /aː/, is changed to ⟨a⟩; the Jyutping ⟨a⟩, which represents /ɐ/, is changed to ⟨e⟩; the Jyutping ⟨e⟩, which represents the nucleus [e] in the final /ei/ and [ɛː], is changed to ⟨ea⟩; the Jyutping ⟨oe⟩, which represents /œː/, is changed to ⟨eo⟩; and the Jyutping ⟨eo⟩, which represents /ɵ/, is changed to ⟨oe⟩. The motivation for each of these modifications is explained in detail below.

**Jyutping ⟨aa⟩ → CantRomZJ1 ⟨a⟩**: This change is motivated by both phonetic correspondence and historical-phonological correspondence. Among Cantonese–Mandarin cognates, Cantonese /aː/ as a nucleus often corresponds to Mandarin /a/, with the two vowels also being phonetically similar [30], [31]. This constitutes a cross-linguistic phoneme correspondence [32]. In MandRomZJ1, the phoneme /a/ is represented by ⟨a⟩. (For symbols that MandRomZJ1 inherits unchanged from Pinyin, we refer only to MandRomZJ1 hereafter.) By contrast, Jyutping represents Cantonese /aː/ as ⟨aa⟩, while reserving the single letter ⟨a⟩ for /ɐ/. Consequently, users familiar with Pinyin or MandRomZJ1 conventions must learn to associate an additional ⟨a⟩ with an /a/-like vowel, while the single ⟨a⟩ represents /ɐ/. To align Cantonese romanization more closely with MandRomZJ1 in terms of both correspondence principles, CantRomZJ1 therefore changes Jyutping ⟨aa⟩ to ⟨a⟩.

**Jyutping ⟨a⟩ → CantRomZJ1 ⟨e⟩**: In the discussion above, Jyutping ⟨aa⟩ is reassigned to CantRomZJ1 ⟨a⟩ for /aː/. Thus, the Jyutping symbol ⟨a⟩ for /ɐ/ must also be reviewed. We choose to use ⟨e⟩ because it improves both correspondence principles with MandRomZJ1 and reduces negative transfer. Phonetically, Cantonese finals /ɐn/ and /ɐŋ/ are close to MandRomZJ1 ⟨en⟩ and ⟨eng⟩ respectively [30]. However, Jyutping ⟨an⟩ and ⟨ang⟩ for /ɐn/ and /ɐŋ/ would make Pinyin or MandRomZJ1 users expect the ⟨an⟩ and ⟨ang⟩ sounds in Pinyin or MandRomZJ1. Similarly, Jyutping ⟨au⟩ for /ɐu/ would make Pinyin or MandRomZJ1 users expect a Pinyin-⟨ao⟩-like or MandRomZJ1-⟨au⟩-like final. Historically, Mandarin has the following cross-linguistic phoneme correspondences with Cantonese /ɐ/-based finals, represented here in MandRomZJ1: Cantonese /ɐi/→MandRomZJ1 ⟨i⟩&⟨uei⟩, /ɐm/→⟨en⟩&⟨in⟩, /ɐn/→⟨en⟩&⟨in⟩, /ɐŋ/→⟨eng⟩&⟨ing⟩, and /ɐu/→⟨ou⟩&⟨iou⟩ [30]. Except for /ɐu/, at least one Mandarin correspondence contains ⟨e⟩. Representing /ɐi, ɐm, ɐn, ɐŋ, ɐu/ as ⟨ei, em, en, eng, eu⟩ can therefore improve historical-phonological correspondence for the first four finals.

**Jyutping ⟨e⟩ → CantRomZJ1 ⟨ea⟩**: In Jyutping, [ɛː] and the nucleus [e] of the final /ei/ are analyzed as one phoneme and represented by ⟨e⟩ [33]. Since the Jyutping ⟨a⟩ has been changed to CantRomZJ1 ⟨e⟩ for /ɐ/, the Jyutping ⟨e⟩ must be reviewed for another romanized form.

Although Mandarin contains the phones [ɛ] and [e] in MandRomZJ1 finals such as ⟨ie⟩, ⟨yue⟩, ⟨ei⟩, and ⟨uei⟩, they are in complementary distribution with [ɤ] and [ə] and are all written as ⟨e⟩ [28]. As in the discussion above, ⟨e⟩ has already been assigned to /ɐ/ in CantRomZJ1. Therefore, MandRomZJ1 does not provide an unused basic Latin letter suitable for representing this Cantonese phoneme. We choose to use ⟨ea⟩ in CantRomZJ1. This choice is motivated by existing romanization usage: ⟨ea⟩ is used to represent [ɛ]-like vowels in the Macau Government Cantonese Romanization and in English, as in bread [34]. Examples in CantRomZJ1 include ⟨sea⟩ for 寫 "write", ⟨neai⟩ for 你 "you", ⟨geang⟩ for 鏡 "mirror", and ⟨ceak⟩ for 尺 "ruler".

This change also helps reduce negative transfer from Pinyin or MandRomZJ1 [9]–[11]. Jyutping finals such as ⟨e⟩ and ⟨eng⟩ represent [ɛː] and [ɛːŋ], whereas ⟨e⟩ and ⟨eng⟩ would lead Pinyin or MandRomZJ1 users to expect [ɤ] or [ə]-like nuclei due to the pronunciations of ⟨e⟩ and ⟨eng⟩ in Pinyin or MandRomZJ1 [28], [33]. Replacing ⟨e⟩ with ⟨ea⟩ therefore reduces the likelihood of such negative transfer for these finals.

This modification nevertheless involves trade-offs. It sacrifices the phonetic correspondence between Cantonese and Mandarin in Chinese characters such as 非, 悲, and 未. These Chinese characters' finals in Cantonese and Mandarin all have /ei/-like pronunciations and are written as ⟨ei⟩ in both Jyutping and MandRomZJ1. However, since Cantonese /ei/ has cross-linguistic phoneme correspondence with a much wider range of Mandarin finals than ⟨ei⟩ alone, the loss in historical-phonological correspondence is relatively limited [30]. The change also sacrifices the historical-phonological correspondence in Chinese characters such as 車, 蛇, and 社, for which both Jyutping and MandRomZJ1 use the final letter ⟨e⟩.

The two modifications discussed above—changing Jyutping ⟨a⟩ to CantRomZJ1 ⟨e⟩ for /ɐ/, and changing Jyutping ⟨e⟩ to CantRomZJ1 ⟨ea⟩ for the nucleus [e] in /ei/ and [ɛː]—have both upsides and downsides. We argue that the upsides are more noticeable and practically beneficial. The upsides include improved historical-phonological correspondence for /ɐi/, /ɐm/, /ɐn/, and /ɐŋ/; improved phonetic correspondence for /ɐn/ and /ɐŋ/; and avoidance of negative transfer caused by Jyutping ⟨an⟩, ⟨ang⟩, ⟨au⟩, ⟨e⟩, and ⟨eng⟩. The downsides include reduced phonetic correspondence for Jyutping ⟨ei⟩ and reduced historical-phonological correspondence for Jyutping ⟨e⟩. This case also demonstrates the balancing consideration between principles.

The proposed modifications are also supported by **statistical evidence** from published language data. In the HKEAA Chinese character list [35], there are 4,159 Chinese characters and 6,907 Cantonese readings. Among the readings, /aː/ occurs in 1,155 readings, while [e] in /ei/ and [ɛː] occur in only 419 readings. Therefore, changing Jyutping ⟨aa⟩ to CantRomZJ1 ⟨a⟩ for the more frequent /aː/, while changing Jyutping ⟨e⟩ to CantRomZJ1 ⟨ea⟩ for the less frequent [ɛː] and [e], produces a net reduction of 1,155 – 419 = 736 keystrokes if all 6,907 readings are typed once using romanized input.

**Jyutping ⟨oe⟩ → CantRomZJ1 ⟨eo⟩**: This modification is mainly motivated by historical-phonological correspondence between Cantonese and neighboring Yue Chinese languages. In many neighboring Yue Chinese languages, Cantonese /œːŋ/ corresponds to /iɔŋ/; representative examples are provided in Supplementary Table S1 [36]. Although the current design framework has not yet been fully extended to non-Cantonese Yue Chinese languages, several ideas suggest that /iɔŋ/ would likely be romanized as ⟨iong⟩ in future extensions. Ideas include Jyutping++ extension practices [8], the current convention of representing /ɔ/ as ⟨o⟩ [33], and our design framework. Changing Jyutping ⟨oeng⟩ to CantRomZJ1 ⟨eong⟩ makes this correspondence clearer: ⟨eong⟩ differs from the expected ⟨iong⟩ by one letter, while Jyutping ⟨oeng⟩ differs by two.

**Jyutping ⟨eo⟩ → CantRomZJ1 ⟨oe⟩**: The Cantonese finals with /ɵ/ as the nucleus include /ɵn/, /ɵy/, and /ɵt/ [33]. Excluding the entering-tone final /ɵt/, /ɵn/ and /ɵy/ show the following cross-linguistic phoneme correspondence with Mandarin finals, represented here in MandRomZJ1 [30]: Cantonese /ɵn/→MandRomZJ1 ⟨in⟩&⟨yun⟩&⟨en⟩&⟨uen⟩ and /ɵy/→⟨uei⟩&⟨yu⟩. CantRomZJ1 changes Jyutping ⟨eon⟩ and ⟨eoi⟩ to ⟨oen⟩ and ⟨oei⟩ respectively. Compared with Jyutping ⟨eon⟩, CantRomZJ1 ⟨oen⟩ shows clearer spelling similarity to MandRomZJ1 ⟨en⟩ and ⟨uen⟩ by sharing the substring ⟨en⟩. Compared with Jyutping ⟨eoi⟩, CantRomZJ1 ⟨oei⟩ also shows such similarity to MandRomZJ1 ⟨uei⟩ by sharing the substring ⟨ei⟩ and differing only in the rounded-vowel letter ⟨o⟩ versus ⟨u⟩. This change therefore improves historical-phonological correspondence at the level of spelling similarity.

### 3.3. Tone Representations

Following the dual-track tone representation, CantRomZJ1 and MandRomZJ1 represent tones with both the category notation for historical-phonological correspondence and the value notation for phonetic correspondence. Table 1 shows the tone names of Cantonese and Mandarin in Chinese linguistics, tone category notations (Cat.), CantRomZJ1 tone value notations (Cant Val.), MandRomZJ1 tone value notations (Mand Val.), Jyutping tone numbers (JP), and Pinyin tone numbers (PY). L, R, D, and E denote level, rising, departing, and entering respectively. Up. and Lo. denote "upper" and "lower" respectively. Unlike Jyutping, which merges entering tones with non-entering tones of the same pitch height [33], CantRomZJ1 keeps them distinct because entering tones differ in syllable length, historical origin, and cross-lingual tone-category correspondence [1], [27].

Table 1: *Tone representations in CantRomZJ1 and MandRomZJ1.*

| Tone Name | Cat. | Cant Val. | Mand Val. | JP | PY |
|---|---|---|---|---|---|
| Yīn L. | 1A | 55 | 55 | 1 | 1 |
| Yáng L. | 1B | 21 | 35 | 4 | 2 |
| R. | 2 | / | 214 | / | 3 |
| Yīn R. | 2A | 35 | / | 2 | / |
| Yáng R. | 2B | 13 | / | 5 | / |
| D. | 3 | / | 51 | / | 4 |
| Yīn D. | 3A | 33 | / | 3 | / |
| Yáng D. | 3B | 22 | / | 6 | / |
| Up. Yīn E. | 4Aa | 5 | / | 1 | / |
| Lo. Yīn E. | 4Ab | 3 | / | 3 | / |
| Yáng E. | 4B | 2 | / | 6 | / |

### 3.4. Justification for Preserving the Jyutping Initial System

The Jyutping initial system is preserved because it aligns well with MandRomZJ1. For the velar, alveolar, and bilabial stop initials, Jyutping symbols provide the unaspirated–aspirated contrast as ⟨b⟩:⟨p⟩, ⟨d⟩:⟨t⟩, and ⟨g⟩:⟨k⟩. For alveolar sibilants, they represent the fricative, unaspirated affricate, and aspirated affricate as ⟨s⟩, ⟨z⟩, and ⟨c⟩, respectively [33]. These conventions are consistent with MandRomZJ1 and reflect both correspondence principles between Cantonese and Mandarin [1]. CantRomZJ1 also preserves ⟨gw⟩ and ⟨kw⟩. Although MandRomZJ1 does not use these as initials, ⟨w⟩ is familiar to Pinyin/MandRomZJ1 users, and treating /w/ as part of the initial maintains the mainstream Cantonese phonology analysis and romanization practice in Hong Kong [28], [33], [34], [37].

A more controversial case might be the Jyutping ⟨j⟩ for /j/. CantRomZJ1 does not change it to MandRomZJ1-like ⟨y⟩. First, we analyze /j/ as an initial rather than as a medial, following the mainstream Cantonese phonology analysis and evidence from diachronic development and native-speaker perception [37]. Second, the issue of ⟨j⟩/⟨y⟩ involves another phoneme whose romanization contains ⟨y⟩: /yː/, for which CantRomZJ1 retains Jyutping ⟨yu⟩. If /j/ were represented as ⟨y⟩, syllables like /jyː/ and /jyːt/ would become ⟨yyu⟩ and ⟨yyut⟩, producing the confusing double-⟨y⟩ sequence. One may consider reducing any double ⟨y⟩ sequence to a single ⟨y⟩ to avoid such confusion. However, this would create further ambiguity: /jyːt/ would be reduced from ⟨yyut⟩ to ⟨yut⟩, sharing the same initial-nucleus substring ⟨yu⟩ with ⟨yuk⟩ representing /jʊk/. Such a representation makes different nuclei share the same romanization.

There are also diachronic and cross-Yue reasons why ⟨j⟩ should be preserved. Reconstructions of Fenyun Cuoyao (分韻撮要), a rime book of the early Qing Cantonese, include an initial /*ɲ/, which later denasalized into modern Cantonese /j/ with the same place of articulation [38]. To reflect this diachronic relationship in a future extension of the romanization design to this early Cantonese, we propose representing /*ɲ/ as ⟨n⟩ plus the modern /j/ symbol. If the modern /j/ were represented as ⟨y⟩, /*ɲ/ would become ⟨ny⟩, and syllables in Fenyun Cuoyao such as /*ɲyː/ would be written as ⟨nyyu⟩, producing another confusing double-⟨y⟩ sequence. Since Fenyun Cuoyao also contains the contrastive syllable /*nyː/ ⟨nyu⟩, the double-⟨y⟩ in ⟨nyyu⟩ cannot be reduced to a single ⟨y⟩ to avoid a collision with ⟨nyu⟩ [38]. Similar contrasts are also found in modern Yue Chinese languages such as Guiping and Baise-Nabi [39], [40]. Thus, preserving ⟨j⟩ is a representative balancing case: although ⟨y⟩ would be superficially closer to MandRomZJ1, using ⟨j⟩ better follows the Cantonese phonological structure, cross-Yue correspondence, and diachronic transparency.

### 3.5. Romanizations for More Sinitic Languages

Although CantRomZJ1 and MandRomZJ1 constitute the main paired case study discussed in detail in this paper, we also apply the design framework to develop romanization schemes for a few other Sinitic languages, including Meixian Hakka, Shanghai Wu, and Nanjing Jianghuai Mandarin. Specific adaptations are allowed when required by the local phonological system. Due to space limitations, the detailed designs of these schemes are not discussed in the main text. Their schemes and design notes are provided in our GitHub repository.

## 4. Open-Source Infrastructure

### 4.1. PhonEngine: System Overview

As discussed in Section 1, existing romanization schemes and pronunciation dictionaries are in heterogeneous formats, making it difficult to develop reusable software. To address this issue, we develop and release PhonEngine, an open-source browser-based HTML system for structured romanization storage and conversion. PhonEngine comprises PhonSymbol, which provides a unified entering interface and structured storage of schemes, and PhonConvert, which supports the efficient large-scale construction of pronunciation dictionaries.

### 4.2. PhonSymbol

PhonSymbol is a lightweight front-end tool with a unified input interface and structured storage of romanization schemes.

The interface supports two editing modes. The first is the medial–nucleus–coda (頭腹尾) mode. This mode is intended for romanization schemes organized into six units: initial, medial, nucleus, coda, syllabic nasal, and tone. Mappings to romanized symbols are defined at the level of these units.

The second mode is the final-based mode. This mode is intended for romanization schemes organized into four units: initial, final, syllabic nasal, and tone. Mappings to romanized symbols are defined at the level of these units.

On the front-end interface, each unit is displayed as an editable grid. Each cell in the editable grid contains two fields: "symbol" and "note". The "symbol" field stores the target romanized symbol, while the "note" field stores the raw symbol or explanatory information. Cells can be added or removed independently in each unit. This design keeps the interface close to the tabular structure familiar to linguists.

PhonSymbol provides a real-time JSON preview of the entered scheme data at the bottom of the interface. Users can export the current scheme as a JSON file or import an existing JSON file to restore and continue editing. The format is simple for manual checking and structured for direct processing by external software. In this study, CantRomZJ1, MandRomZJ1, and the additional Sinitic romanization schemes are encoded using this JSON-based PhonSymbol format.

The interface screenshots for these two modes are shown in Supplementary Figure S1.

### 4.3. PhonConvert

The practical use of a romanization scheme requires pronunciation dictionaries with sufficient Chinese character coverage, but manually annotating each Chinese character is costly and difficult to scale across multiple languages. PhonConvert is therefore designed as a browser-based front-end tool for rapidly constructing large-scale romanized pronunciation dictionaries from existing language resources.

PhonConvert currently supports two input modes. The first mode is designed for Kaom.net, one of the largest open platforms for Sinitic language pronunciation data covering more than 2,000 language locations [41]. In this mode, users can copy the content of a target language pronunciation page and paste it into the input area. The second mode accepts CSV-style character-pronunciation data, where each row contains Chinese characters, an initial, a final, and a tone. In both modes, the input data are extracted into the same internal structure and passed to the same phonology-and-romanization organizer workflow.

A guide to using PhonConvert with Kaom.net as a data source is provided in Supplementary Figures S2–S7.

After data extraction, users can enter the phonology-and-romanization organizer, a visual workspace for converting the parsed raw data to target romanized syllables (Figure 1). The top bar, marked in yellow, displays all parsed raw final symbols and allows users to select a final tab. The red region shows, for each row, the parsed raw initial symbol, the parsed raw tone symbols that form legal syllables with the selected final, and the Chinese characters with the corresponding pronunciation. These parsed data and Chinese characters are all in editable fields. The raw final symbols in the top bar cannot be directly edited there, but changes made in the "Final IPA/RAW" field are reflected in the final tabs. The orange regions are used to enter mapping information from parsed raw symbols to target romanized symbols together with notes. The orange subregions numbered 1, 2, and 3 contain initial, final, and tone mapping information, respectively. For initial mapping, the target romanized initial symbol in the General field applies to the initial for all finals, while the target romanized initial symbol in the Specific field only applies to the initial for the currently selected final. If the target romanized initial symbol in the Specific field has the form -A+B and the selected final starts with A in its romanization, the leading A in the final will be replaced by B. If the selected final has the romanized form ⟨ian⟩ and the target romanized initial symbol in the Specific field is ⟨-i+y⟩, the romanized syllable has the initial-final form ⟨yan⟩. This design allows romanized syllables with the Pinyin/MandRomZJ1 style to be generated. For tone mapping, the parsed raw tone symbols in the red region are parsed from the source data, while those in the orange region are user-defined keys mapping to the target romanized tone symbols. During syllable generation, a parsed raw tone symbol in the red region receives a target romanized symbol only when the parsed symbol matches a user-defined key in the orange region. Finally, the indigo region displays the generated romanized syllables, which are composed of the romanizations of the currently selected final, each row's initial, and the corresponding tone in that row.

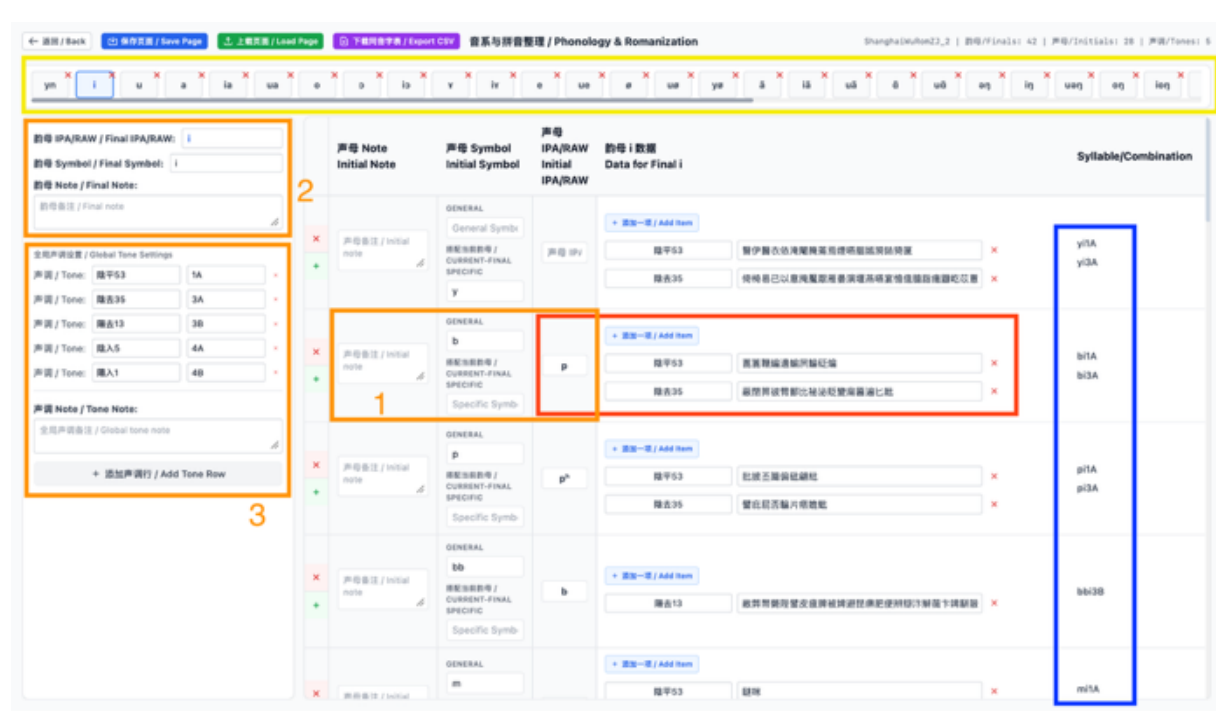

Figure 1: *PhonConvert phonology-and-romanization organizer. A larger, clearer version is provided in Supplementary Figure S8.*

PhonConvert provides two output functions. First, users can save the current editing state as a JSON file, which can later be reloaded to restore and continue the work. Second, users can export the finalized data as a CSV homophone-character table in the format "romanized syllable,character list", such as "ssen1B,人神仍" for a Shanghai Wu example. This format follows the traditional homophone-character table practice in Chinese linguistics while remaining easy for software to process.

In this study, we use PhonConvert to generate large-scale pronunciation dictionaries for CantRomZJ1, MandRomZJ1, and the additional Sinitic romanization schemes introduced above. For each scheme, we also provide the final editing state data as a JSON file, making the conversion process transparent, reproducible, and available for further editing.

### 4.4. Open-Source Resources for CantRomZJ1

We also release open-source data and tools for CantRomZJ1. The Jyutping Workgroup of LSHK has released the Cantonese Pronunciation List of the Characters for Computers, which provides Jyutping pronunciations for more than 29,000 Chinese characters in JSON and TSV formats [18]. Based on this resource, we convert the Jyutping data in these files into CantRomZJ1 while preserving the original file structures where possible to support compatibility with existing scripts and workflows.

To connect CantRomZJ1 with the existing Jyutping NLP ecosystem, we provide PyCantonese-compatible conversion tools, including scripts for converting PyCantonese-style Jyutping strings to CantRomZJ1, converting CantRomZJ1 strings to Jyutping, and parsing CantRomZJ1 syllables into the same structure used by PyCantonese's Jyutping syllable parser [16]. The same functions are also released through PyPI as part of PySinRom. Together, these resources allow CantRomZJ1 to reuse existing Jyutping data and software.

### 4.5. Conversion to Rime YAML Files

To connect PhonConvert outputs with practical input-method deployment, we provide tools for converting the CSV homophone-character tables exported by PhonConvert to Rime-compatible YAML files [15], [20]. This CSV-to-Rime conversion is available both as a Python script and as a browser-based HTML front-end for users who do not use Python.

## 5. S2R Validation of Proposed Schemes

To provide preliminary computational validation of the design framework, we conduct a Mandarin–Cantonese joint S2R experimental comparison by fine-tuning MMS [21]. We compare two target-representation conditions. The baseline condition, named pyjp, uses Pinyin as the Mandarin prediction target and Jyutping as the Cantonese prediction target. The design framework condition, named romzj1, uses MandRomZJ1 as the Mandarin prediction target and CantRomZJ1 as the Cantonese prediction target. Although Cantonese S2R is the primary evaluation task, both conditions are jointly trained on Mandarin and Cantonese to examine whether the MandRomZJ1–CantRomZJ1 pair can facilitate transfer from higher-resource Mandarin to lower-resource Cantonese. The two conditions use identical audio data, data splits, model architecture, training procedure, and hyperparameters. The target romanization representation is the only experimental factor that differs between them.

### 5.1. Data and Targets

To simulate the low-resource setting, we sample approximately 1 hour of Cantonese speech and 10 hours of Mandarin speech as the training set, with additional validation and test sets corresponding to about 10% and 20% of the training size respectively. The Mandarin data are sampled from AISHELL-3 [42], whose studio-quality recordings help reduce noise confounds in the S2R experiment. The Cantonese data are sampled from MDCC [43]. Since part of MDCC contains background music, we first filter the dataset to retain clean speech before sampling. The sampled file names in both datasets are provided in our GitHub repository.

Romanization targets are generated from the Chinese-character transcripts using Python packages pypinyin and ToJyutping [17], [44]. Although AISHELL-3 provides human-labeled Pinyin transcriptions, these labels faithfully reflect surface speech realization, including tone sandhi and erhua-like forms such as ⟨miaor⟩ for 廟 without a following 兒. Such forms are not standard dictionary entries and are less suitable for the dictionary-driven P2G stage in a practical cascading ASR system. Therefore, to better approximate practical S2R-to-P2G usage, we use pypinyin and ToJyutping to generate standard dictionary form-based romanization targets. MandRomZJ1 and CantRomZJ1 targets are obtained from Pinyin and Jyutping targets, respectively, using deterministic scripts.

During data inspection, we found a systematic ToJyutping labeling error involving the character 呢: single-character tokens not segmented as part of a multi-character word were labeled as ⟨ne1⟩. Native-speaker verification indicated that non-sentence-final 呢 should be labeled as ⟨ni1⟩ rather than ⟨ne1⟩. We therefore automatically corrected all non-sentence-final 呢 labels to ⟨ni1⟩ in both the Jyutping and CantRomZJ1 targets.

Each target sequence begins with a language-identification token, following common practice in multilingual language technology [45]. The remaining romanization string is serialized as a sequence of individual characters—Latin letters and tone digits—with adjacent characters separated by a delimiter token. The delimiter is included in the CTC target sequence and is decoded as whitespace. This serialization allows each character to function as an individual prediction token and, after decoding, as a whitespace-delimited token for jiwer-based WER computation [46]. In the WER computation, whitespace serves as a token boundary and is not counted as an independent evaluation unit, whereas CER is computed over the decoded character string and therefore includes whitespace characters.

### 5.2. Model and Training Configuration

We fine-tune facebook/mms-300m as a Wav2Vec2ForCTC model with a CTC objective [21], [47], [48]. The two target-representation conditions use the same model architecture, while the output vocabulary and CTC prediction layer are constructed separately for their respective target representations. We freeze only the CNN feature encoder and fine-tune all remaining parameters.

Each target-representation condition is evaluated under two target-label settings: one with the systematic 呢 labeling error corrected and one without the correction. For each combination of target-representation condition and target-label setting, we conduct three random-seed runs using seeds 41, 42, and 43. This results in 12 joint-training runs for the pyjp–romzj1 comparison. The main text reports the six runs under the corrected target-label setting, while the corresponding six runs without the correction are reported in Supplementary Table S2. All runs use identical data splits, training procedures, and hyperparameters. Detailed training hyperparameters are provided in Supplementary Table S3.

### 5.3. Pilot Comparison of Tone Encoding

Before the main pyjp versus romzj1 comparison, we conduct a pilot comparison of tone encodings for the ASR target representation. As discussed in Section 2, tone differs from initials and finals because it is a suprasegmental feature and may need digits other than Latin letters [23]. The HTCN is linguistically informative because it encodes historical origins explicitly. However, when treated as character-level CTC targets, this notation makes each tone correspond to multiple tokens, lengthens the target sequence, and may make alignment with audio frames more difficult [47], compared to the 1–6 and 1–4 tone notations in Jyutping and Pinyin [1], [31], [33], [44].

We therefore compare three ASR tone encodings. The first uses the HTCN as character-level targets. For Mandarin, the second and third both use a 1–4 single-token encoding corresponding to yīn level, yáng level, rising, and departing. For Cantonese, the second uses nine single-token labels, a 1–9 encoding, corresponding to yīn level, yáng level, yīn rising, yáng rising, yīn departing, yáng departing, upper yīn entering, lower yīn entering, and yáng entering [33]. This follows a traditional Chinese linguistics tonal ordering convention and makes 1 and 2 correspond to yīn level and yáng level in both Mandarin and Cantonese [44], [49]. For Cantonese, the third uses a compact 1–6 single-token encoding corresponding to yīn level, yáng level, yīn rising, yáng rising, yīn departing, and yáng departing, while the three entering tones use the same digits as non-entering tones with the same pitch height.

We do not use a tone-value target here because Mandarin neutral tone has no fixed tone value. Assigning context-dependent values according to the preceding tone would introduce partial surface speech phonetic transcription into otherwise dictionary form-based romanization targets [50], which may create inconsistencies and noise for downstream P2G.

The pilot results indicate that the third encoding performs better than both the first and second encodings. Using the romzj1 seed-42 setting, the third encoding achieved Cantonese WER/CER of 0.0692/0.0429, compared with 0.0882/0.0565 for the second one and 0.1667/0.1112 for the first one. We interpret this result as follows: the first encoding lengthens the CTC target sequence and degrades performance as discussed in the first paragraph. The second encoding increases label sparsity for entering tones, and the model not only needs to distinguish the entering tone pitch height but also to recognize their checkedness to predict them as three independent tokens. With the third encoding, entering tones share tokens with non-entering tones of the same pitch height, so the model can predict tone mainly from pitch information. We therefore use the third encoding in all main S2R experiments. This is an S2R-specific technical adaptation and does not replace the first HTCN encoding proposed above, which is still linguistically reasonable.

### 5.4. Results and Discussion

Table 2 reports the Cantonese test results under the corrected target condition. Across all three random seeds, romzj1 consistently outperforms the pyjp baseline. On average, MandRomZJ1+CantRomZJ1 reduce Cantonese WER from 0.0769 to 0.0709 and CER from 0.0490 to 0.0438, corresponding to relative reductions of 7.80% and 10.61%, respectively.

Table 2: *Cantonese test results with the corrected 呢 target.*

| Seed | pyjp WER | pyjp CER | romzj1 WER | romzj1 CER |
|---|---|---|---|---|
| 41 | 0.0845 | 0.0581 | 0.0717 | 0.0440 |
| 42 | 0.0724 | 0.0442 | 0.0692 | 0.0429 |
| 43 | 0.0738 | 0.0446 | 0.0719 | 0.0444 |
| Mean | 0.0769 | 0.0490 | 0.0709 | 0.0438 |

The uncorrected automatic-target results are provided in Supplementary Table S2. The same trend is observed: romzj1 outperforms pyjp across all seeds, suggesting that the advantage is not merely an artifact of the rule-based 呢 correction.

To distinguish cross-lingual transfer from differences in the intrinsic difficulty of the Cantonese target representations, we additionally conduct a Cantonese-only comparison under two target-representation conditions: one using Jyutping and the other using CantRomZJ1. Both conditions use the corrected 呢 targets. For each condition, we conduct three random-seed runs using seeds 41, 42, and 43, resulting in six Cantonese-only runs. As shown in Table 3, CantRomZJ1 reduces mean WER by only approximately 1.9% and mean CER by approximately 0.2% relative to Jyutping. These reductions are substantially smaller than the 7.80% WER and 10.61% CER reductions observed under joint Mandarin–Cantonese training. Moreover, with random seed 41, Jyutping outperforms CantRomZJ1 in both WER and CER. The results therefore indicate that Jyutping and CantRomZJ1 have broadly comparable difficulty under Cantonese-only training. The larger gains under joint training are more plausibly associated with cross-lingual romanization alignment than with CantRomZJ1 being intrinsically easier. Seed-level results are provided in Supplementary Table S4.

Table 3: *Mean Cantonese-only results across three seeds.*

| Cant. target | WER | CER | Rel.WER Reduc. | Rel. CER Reduc. |
|---|---|---|---|---|
| Jyutping | 0.0829 ± 0.0023 | 0.0545 ± 0.0016 | / | / |
| CantRomZJ1 | 0.0813 ± 0.0042 | 0.0544 ± 0.0050 | 1.9% | 0.2% |

These results support the central computational claim of this paper. Since the two conditions use identical audio data, model architecture, training method, and hyperparameters, the performance difference is more plausibly associated with the cross-lingual alignment of the target romanization representations. These results are consistent with the proposed design framework: by improving correspondence between Mandarin and Cantonese romanizations, MandRomZJ1+CantRomZJ1 may help a shared model reuse acoustic-to-symbol mappings for lower-resource Cantonese. The experiment also shows that tone design is sensitive to the target task. The HTCN explicitly represents historical-phonological information, but it is not necessarily optimal as a CTC target representation [47]. In this CTC-based S2R experiment, the more compact 1–6 tone representation performs better.

## 6. Conclusion

In this paper, we propose the Sinitic Romanization Ecosystem, comprising a cross-lingual romanization design framework, supporting digital infrastructure, and a community-driven open-source workflow. The design framework includes four principles: phonetic correspondence, historical-phonological correspondence, one-phoneme-one-symbol, and basic Latin-letter use, with a balancing consideration recognizing trade-offs among these principles. MandRomZJ1 and CantRomZJ1, as new Mandarin and Cantonese romanization schemes respectively, serve as the main paired case study. A limitation of the present study is that it does not directly evaluate how romanizations under the design framework affect human learning outcomes and language-technology tasks beyond ASR, such as TTS. The design framework may also be applicable to other Asian languages that share relevant properties, including Sinitic lexical borrowing, tonal or syllable-based structures, and contact-induced phonological correspondences. Future work will evaluate its effects on human learning and additional language-technology tasks and explore schemes beyond Sinitic languages.

# Supplementary Material

# Toward a Cross-Lingual Romanization Ecosystem for Sinitic Languages: A Paired Mandarin–Cantonese Case Study

*Zijie Zhang, Tan Lee, Yong Cao, Benyou Wang*

## S1. PhonEngine Interfaces

Figure S1 shows the two PhonSymbol editing modes described in Section 4.2 of the main paper. The final-based mode stores initial, final, syllabic-nasal, and tone mappings, whereas the medial–nucleus–coda mode stores initial, medial, nucleus, coda, syllabic-nasal, and tone mappings.

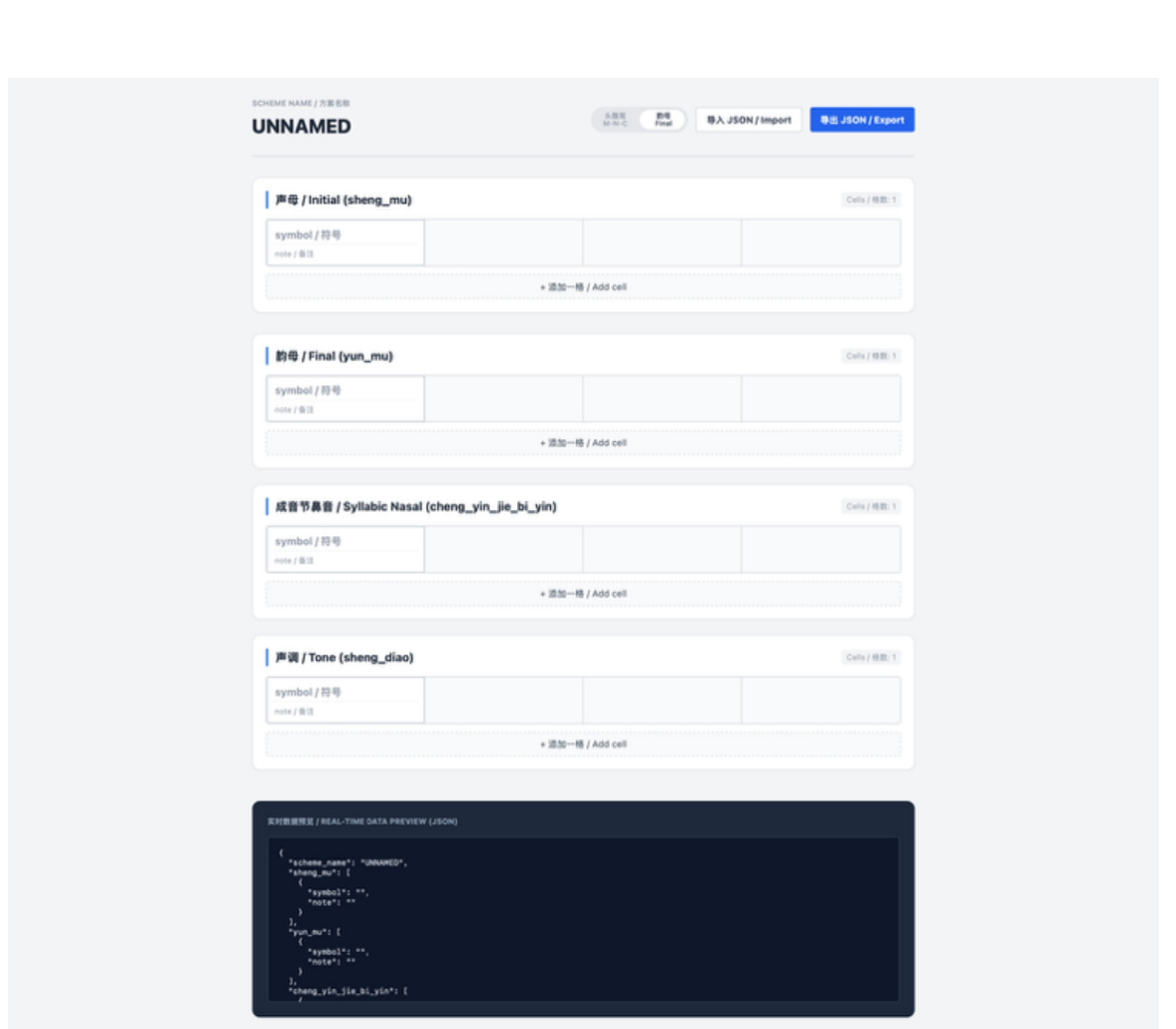

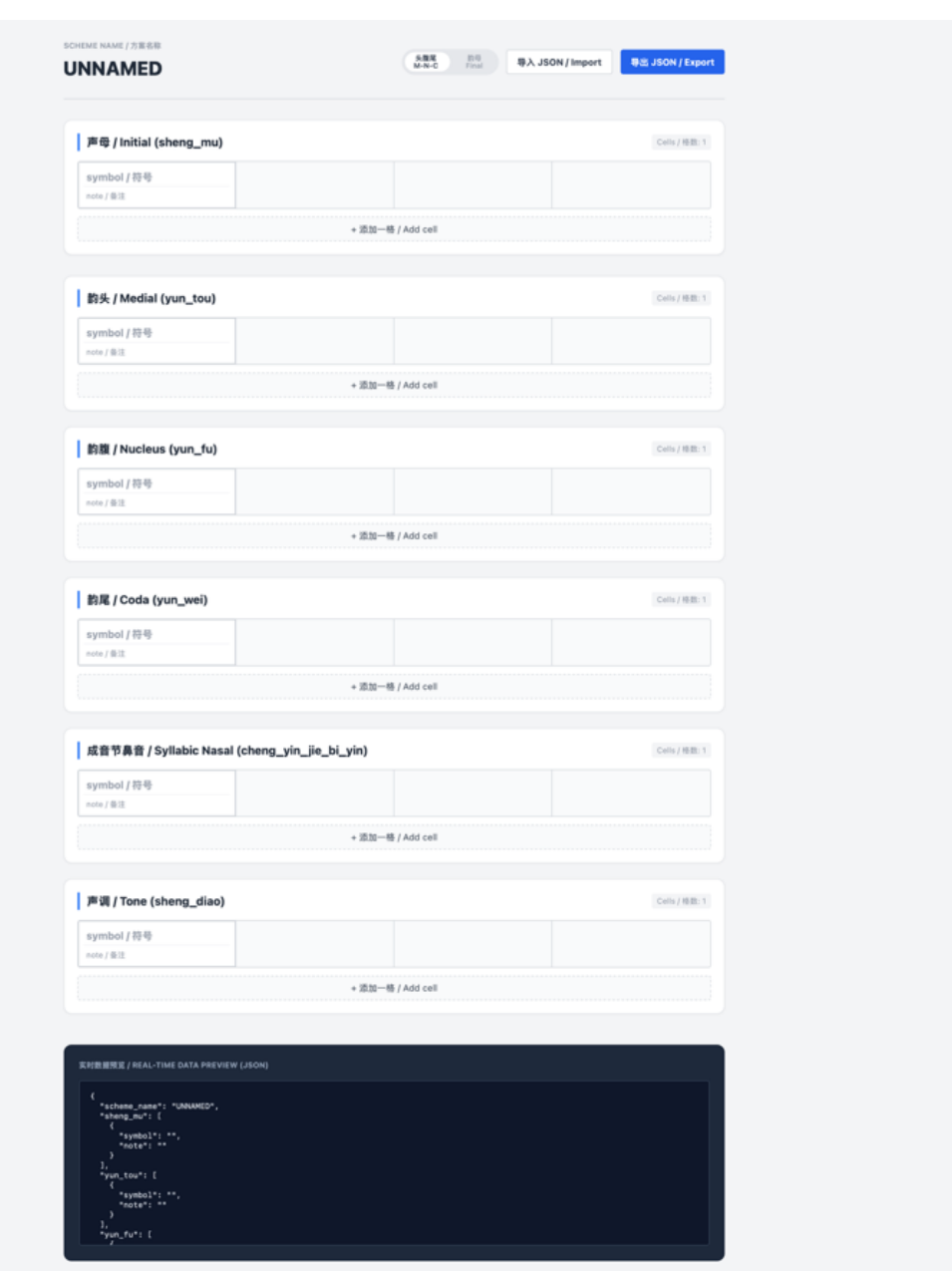

Figure S1: *PhonSymbol interfaces. Left: final-based mode. Right: medial–nucleus–coda mode.*

## S2. Using PhonConvert with Kaom.net as a Data Source

The following workflow illustrates how a Kaom.net pronunciation page can be converted into structured romanization data with PhonConvert.

**Step 1.** Open the Kaom.net Sinitic-language pronunciation index at http://www.kaom.net/si_x.php (Figure S2) and select a target language location.

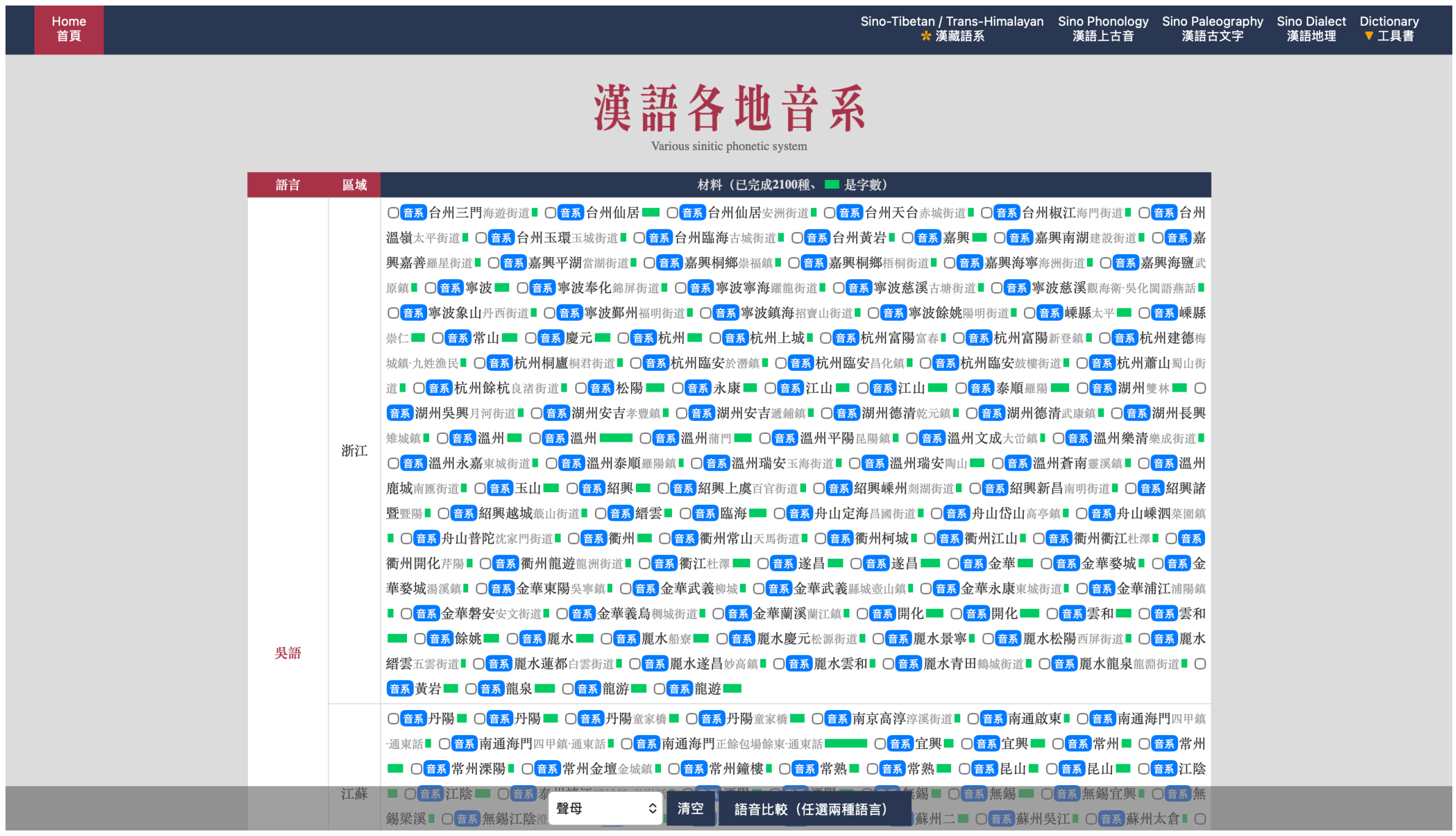

Figure S2: *Kaom.net page listing supported Sinitic-language locations.*

**Step 2.** Click the “音系” (phonology) button in Figure S2 for the selected location to open its phonology page (Figure S3). From Figure S3, click “全字表” (full character table) to open the character-level pronunciation table (Figure S4).

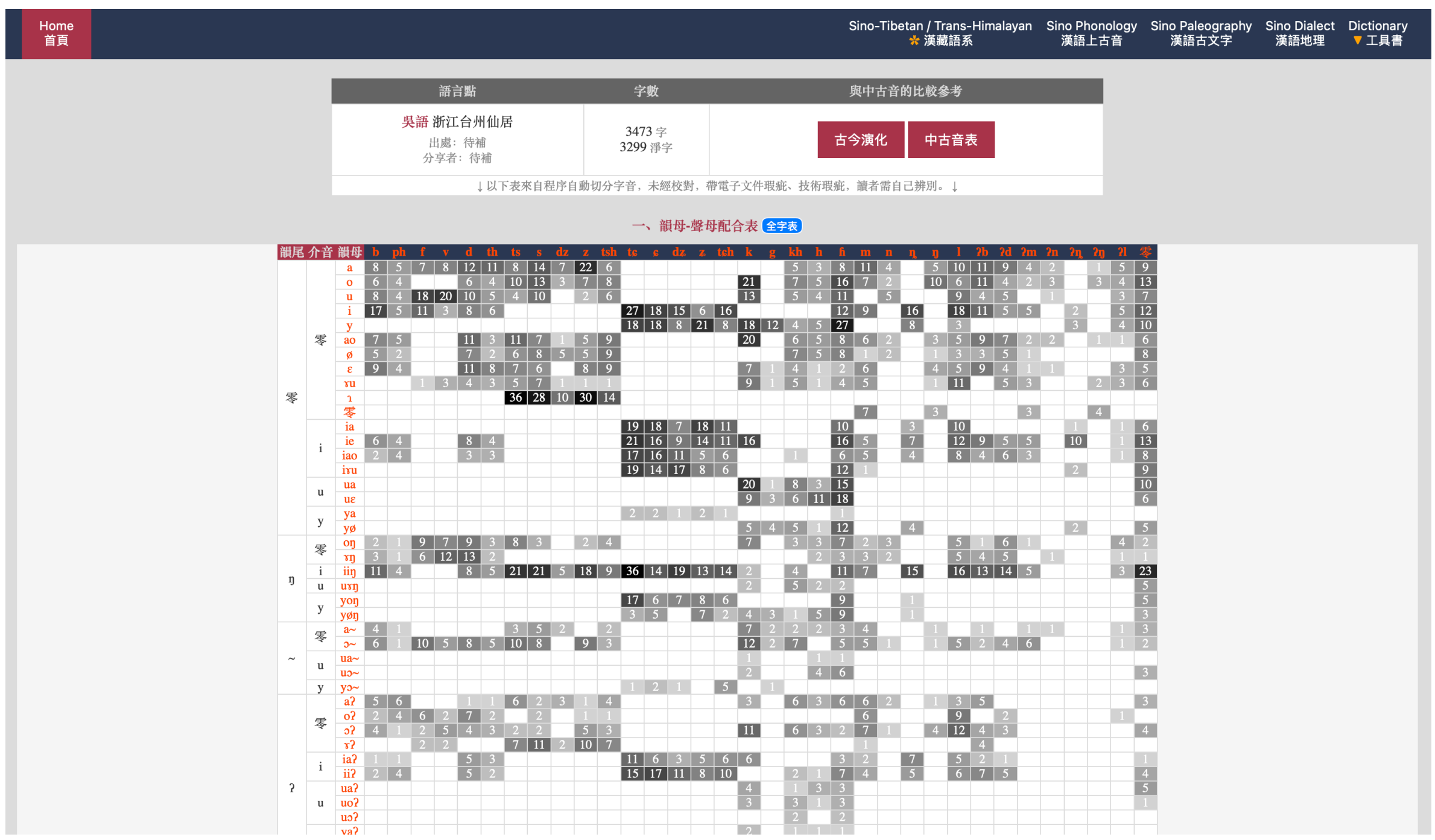

Figure S3: *Kaom.net phonology page after selecting a language location.*

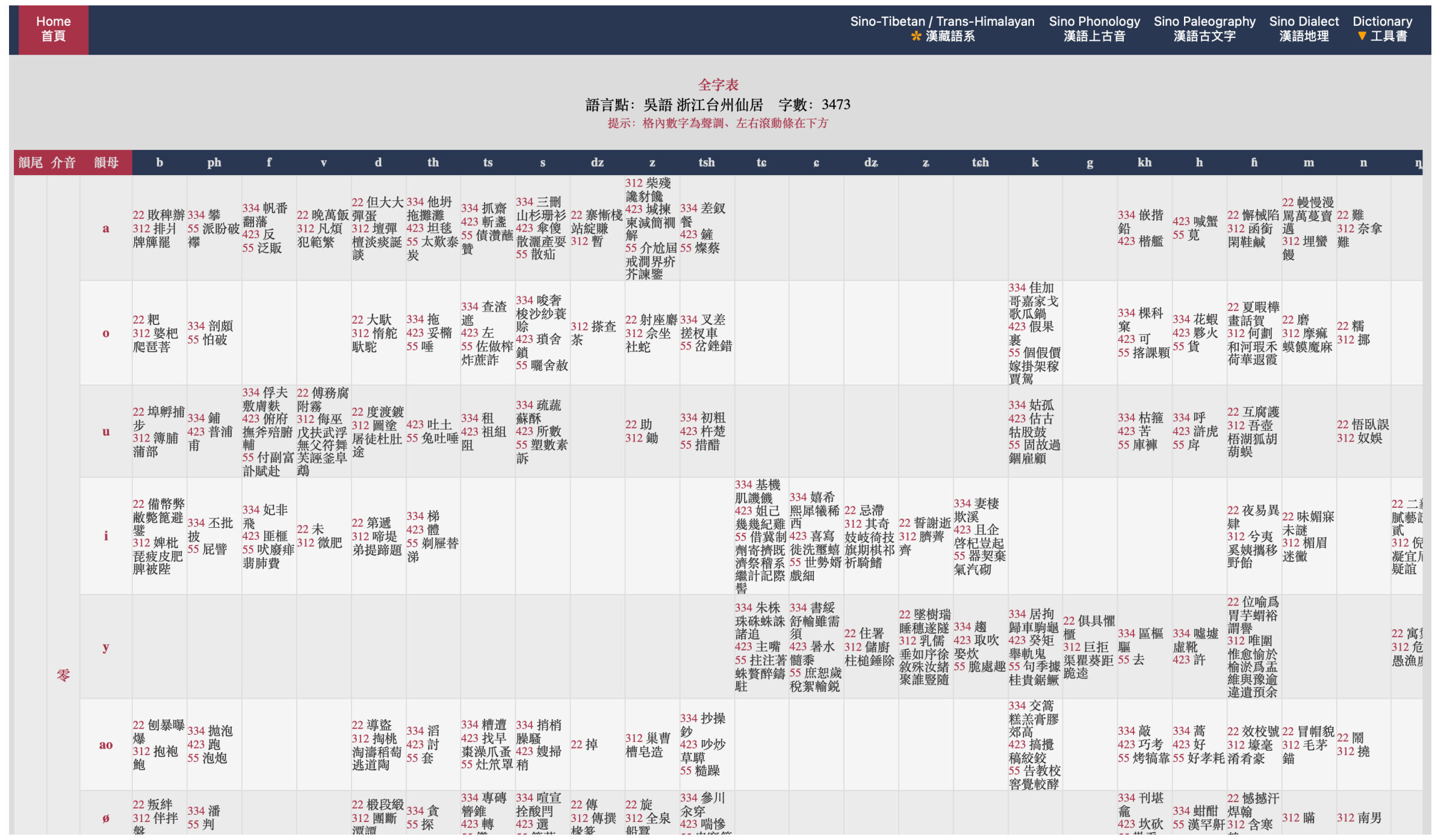

Figure S4: *Kaom.net full character-pronunciation table for the selected language location.*

**Step 3.** On the full character-pronunciation table (Figure S4), select all page content (Ctrl+A on Windows/Linux or Command+A on macOS), copy it, and paste it into the Rich Text Editor on the first page of PhonConvert (Figure S5).

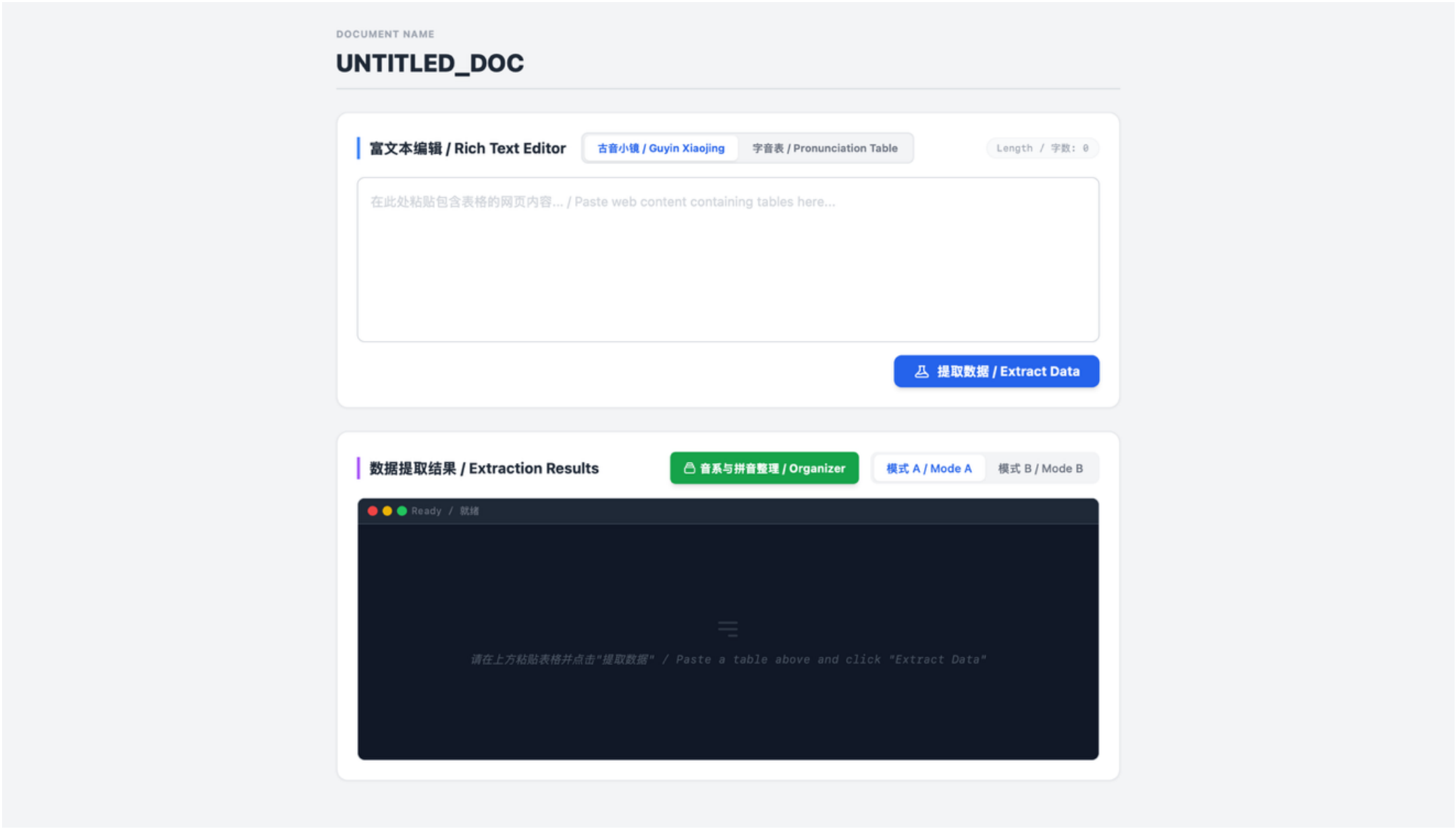

Figure S5: *PhonConvert input page. Kaom.net rich text can be pasted into the Rich Text Editor before data extraction.*

**Step 4.** Click "Extract Data" and then "Organizer" on Figure S5 to enter the phonology-and-romanization organizer (Figure S6), where parsed source symbols can be mapped to target romanized symbols and generated syllables can be inspected and edited.

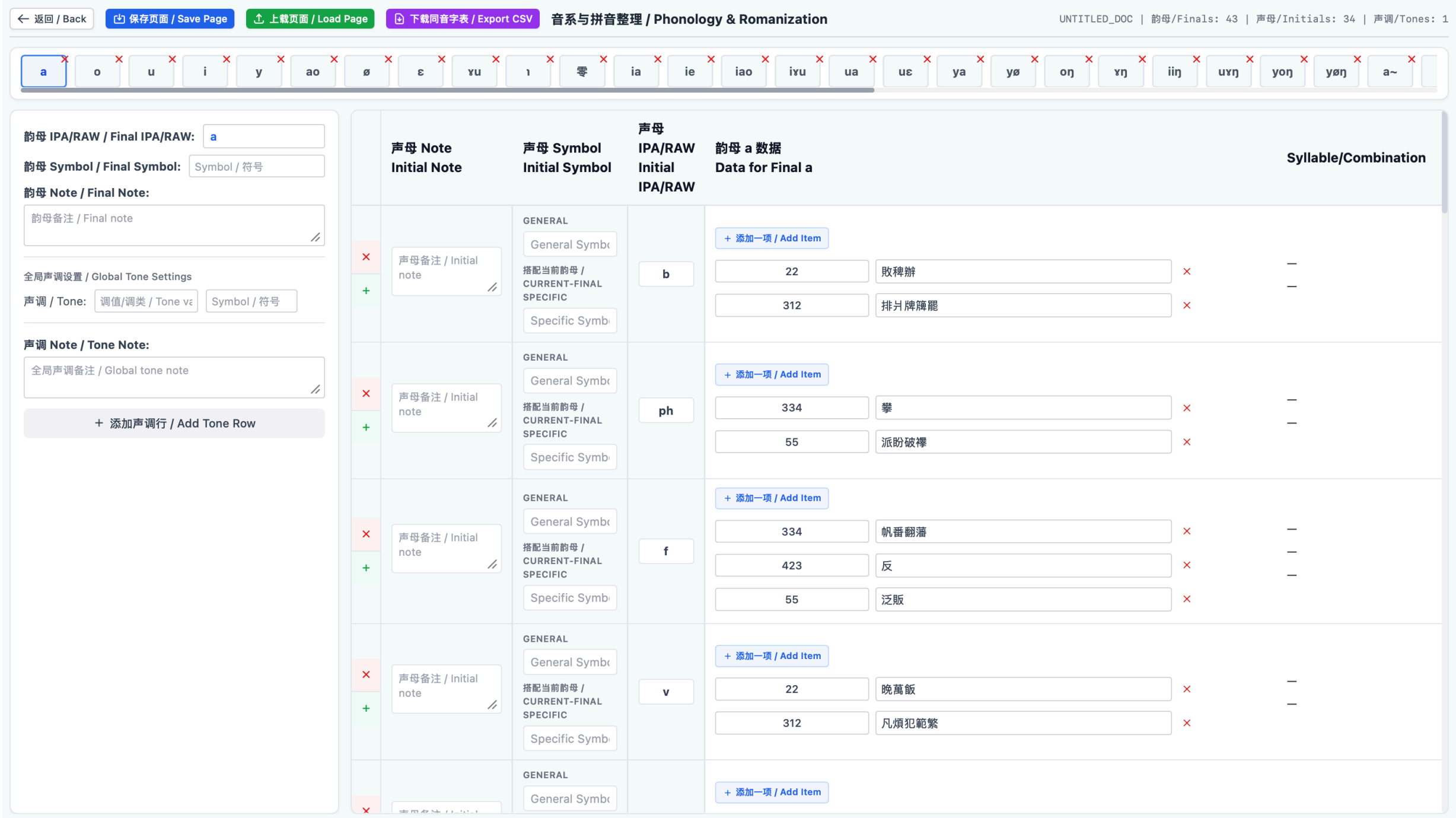

Figure S6: *PhonConvert phonology-and-romanization organizer after parsing a Kaom.net source page.*

**Step 5.** PhonConvert can also be used without importing Kaom.net data. From the blank first page of PhonConvert (Figure S5), click “Organizer” to enter a blank phonology-and-romanization organizer (Figure S7) to work from blank or upload a previously saved editing-state JSON file to restore and continue an existing project.

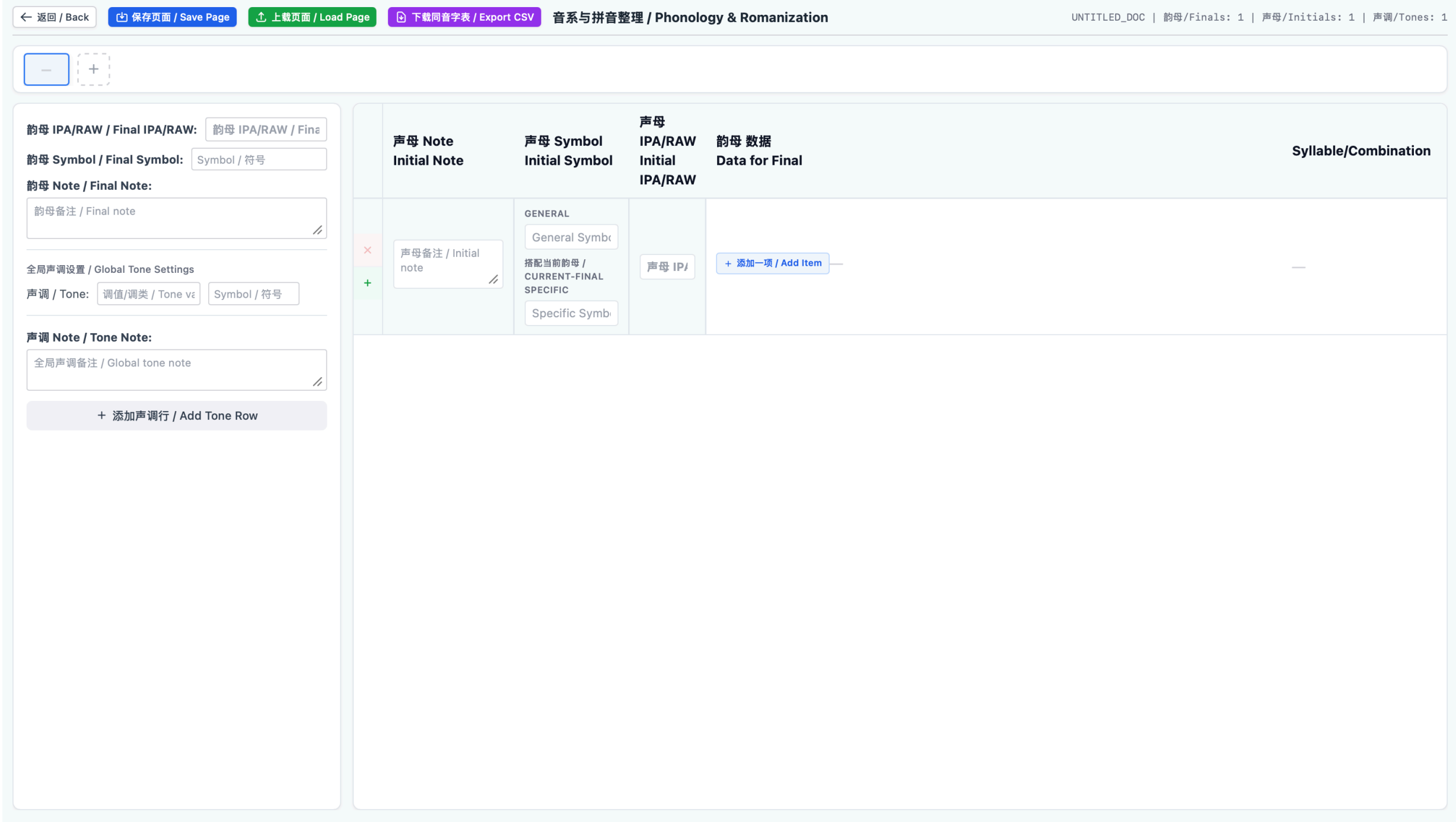

Figure S7: *Blank PhonConvert phonology-and-romanization organizer.*

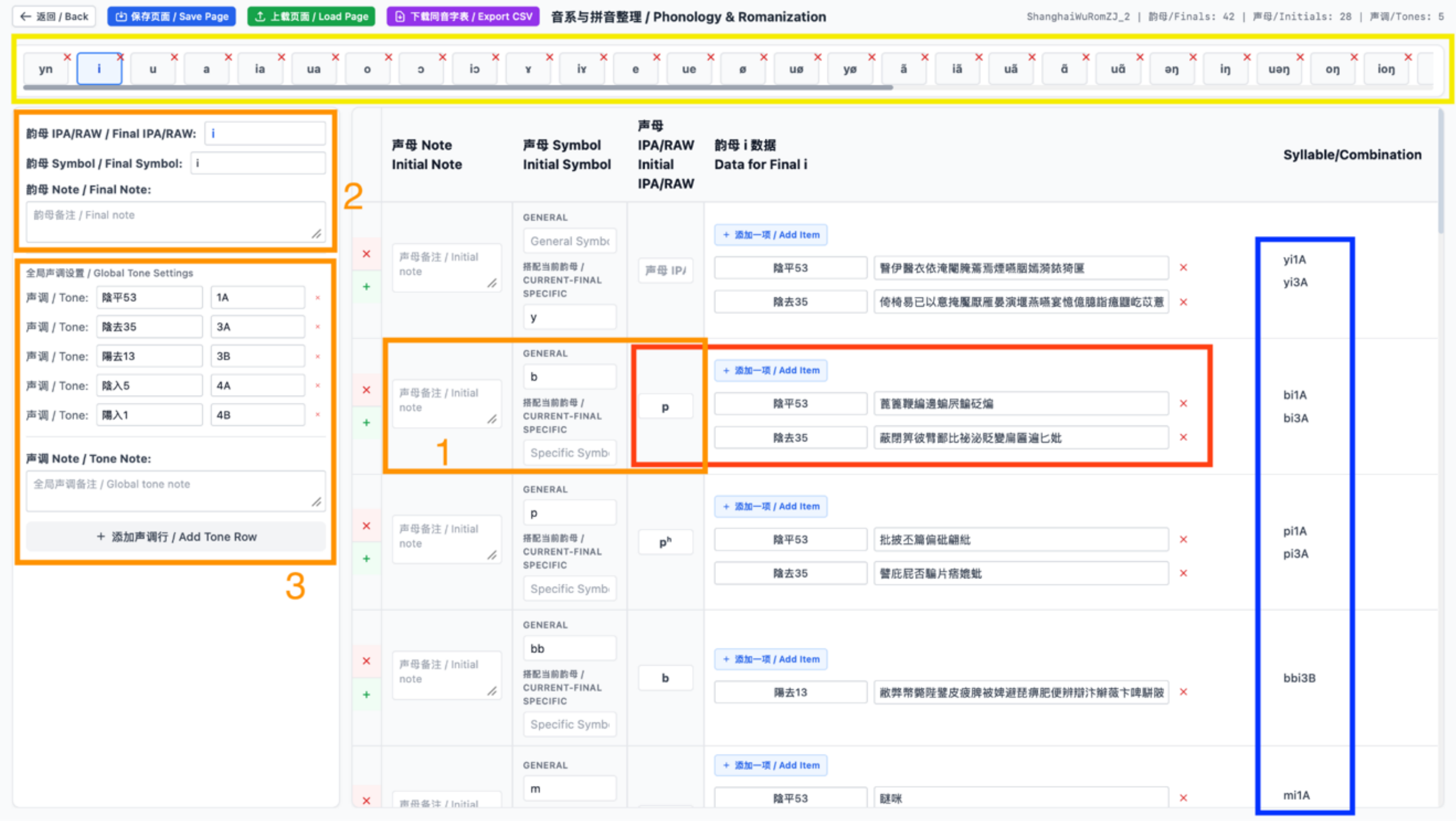

Figure S8: *Annotated, higher-resolution view of the PhonConvert phonology-and-romanization organizer.*

***Note.*** *The organizer shown in Figure S8 uses a different Sinitic-language example from the workflow in Figures S2–S7.*

## S3. Additional Phonological Evidence

Table S1: *Representative examples of Cantonese /œːŋ/ corresponding to /iɔŋ/ in selected neighboring Yue Chinese languages.*

| Finals | 想 | 香 | 將 | 梁 | 釀 |
|---|---|---|---|---|---|
| Guangzhou | œːŋ | œːŋ | œːŋ | œːŋ | œːŋ |
| Huaxian (Huashan) | iɔŋ | iɔŋ | iɔŋ | iɔŋ | iɔŋ |
| Sanshui (Xi'nan) | iɔŋ | iɔŋ | iɔŋ | iɔŋ | iɔŋ |
| Xinhui (Huicheng) | iɔŋ | iɔŋ | iɔŋ | iɔŋ | iɔŋ |
| Doumen Zhen | iɔŋ | iɔŋ | iɔŋ | iɔŋ | iɔŋ |
| Enping (Niujiang) | iɔŋ | iɔŋ | iɔŋ | iɔŋ | iɔŋ |
| Baoan (Shajing) | iɔŋ | iɔŋ | iɔŋ | iɔŋ | iɔŋ |

***Note.*** *The table lists representative Chinese characters for which Cantonese /œːŋ/ corresponds to /iɔŋ/ in selected neighboring Yue Chinese languages. The data are based on Xiaoxuetang Yue [36]. The table illustrates the historical-phonological motivation for changing Jyutping ⟨oeng⟩ to CantRomZJ1 ⟨eong⟩ and is not intended as a complete comparative survey of Yue reflexes.*

## S4. Additional S2R Results and Training Details

Table S2: *Cantonese S2R test results under the uncorrected 呢 target condition.*

| Seed | pyjp WER | pyjp CER | romzj1 WER | romzj1 CER |
|---|---|---|---|---|
| 41 | 0.0783 | 0.0498 | 0.0761 | 0.0477 |
| 42 | 0.0742 | 0.0495 | 0.0714 | 0.0446 |
| 43 | 0.0731 | 0.0506 | 0.0704 | 0.0424 |
| **Mean** | 0.0752 | 0.0500 | 0.0726 | 0.0449 |

Table S3: *Training hyperparameters for MMS fine-tuning.*

| Setting | Value |
|---|---|
| Model | facebook/mms-300m |
| Objective | CTC |
| Frozen module | CNN feature encoder |
| Per-device batch size | 8 |
| Gradient accumulation | 4 |
| Effective batch size | 32 |
| Maximum epochs | 30 |
| Learning rate | 3e-4 |
| Warmup steps | 1,000 |
| Evaluation interval | 100 steps |
| Checkpoint interval | 100 steps |
| Checkpoint selection | Best validation WER |
| Random seeds | 41, 42, 43 |

Table S4: *Cantonese-only S2R results across three random seeds.*

| Seed | Jyutping WER | Jyutping CER | CantRomZJ1 WER | CantRomZJ1 CER |
|---|---|---|---|---|
| 41 | 0.0805 | 0.0539 | 0.0842 | 0.0591 |
| 42 | 0.0851 | 0.0563 | 0.0833 | 0.0549 |
| 43 | 0.0831 | 0.0533 | 0.0765 | 0.0491 |